\documentclass[graybox]{svmult}

\usepackage{mathptmx}       
\usepackage{helvet}         
\usepackage{courier}        
\usepackage{type1cm}        
\usepackage{makeidx}         
\usepackage{graphicx}        
\usepackage{multicol}        
\usepackage[bottom]{footmisc}

\makeindex             

\begin{document}

\title*{Understanding How People Rate Their Conversations}
\author{Alexandros Papangelis, Nicole Chartier, Pankaj Rajan, Julia Hirschberg, Dilek Hakkani-Tur}
\institute{Alexandros Papangelis \at Amazon Alexa AI, \email{papangea@amazon.com}
}
%
%
\maketitle

\abstract{User ratings play a significant role in spoken dialogue systems. Typically, such ratings tend to be averaged across all users and then utilized as feedback to improve the system or personalize its behavior. While this method can be useful to understand broad, general issues with the system and its behavior, it does not take into account differences between users that affect their ratings. In this work, we conduct a study to better understand how people rate their interactions with conversational agents. One macro-level characteristic that has been shown to correlate with how people perceive their inter-personal communication is personality \cite{astridKramerGratch10, cupermanIckes09, mccraeCosta89}. We specifically focus on agreeableness and extraversion as variables that may explain variation in ratings and therefore provide a more meaningful signal for training or personalization. In order to elicit those personality traits during an interaction with a conversational agent, we designed and validated a fictional story, grounded in prior work in psychology. We then implemented the story into an experimental conversational agent that allowed users to opt-in to hearing the story. Our results suggest that for human-conversational agent interactions, extraversion may play a role in user ratings, but more data is needed to determine if the relationship is significant. Agreeableness, on the other hand, plays a statistically significant role in conversation ratings:  users who are more agreeable are more likely to provide a higher rating for their interaction. In addition, we found that users who opted to hear the story were, in general, more likely to rate their conversational experience higher than those who did not.}

\section{Introduction}
User feedback is one of the most important pieces of information we can use to improve various modules of conversational agents. Such feedback is usually provided by directly asking users to rate their experience (e.g. on a scale of 1 to 5). These ratings are typically averaged and used as a measure of the agent's conversational skills. 

One limitation of this approach is that it treats all users as an homogeneous whole. However, each user is different; they have different experiences, personalities, needs, and expectations that can lead them to perceive an interaction with the same conversational agent differently. Treating conversational ratings as monolithic will lead to a conversational agent that tends to an `average' user, rather than being personalized to each individual user. While for some users, an `averaged' approach will not have an adverse effect on the conversational experience, this approach may lead to sociodemographic and personality bias in the agent, and negative experiences for some users. 

Because of this, we postulate that learning about users, and using that information to personalize a conversational agent, will improve the user's conversational experience and thus, improve their rating of the conversation. In this work, we propose approaching conversational ratings through the lens of users' personality to address the question \textit{Does a user's personality play a role in the rating they provide?}

We chose to focus on personality for two reasons. First, personality is a well-documented and researched area in which individual variation can be explained using macro-level categorization \cite{goldberg2006ipip, goldberg1992bigfive, goldberg99, john1999big}. As such, methods for measuring individuals' personality traits via surveys have been thoroughly assessed and validated. Second, previous research suggests that two personality traits, extraversion and agreeableness, influence a user's evaluation of their interaction with conversational agents \cite{astridKramerGratch10}. To gain some insight into users' personalities, we constructed a story that includes questions about agreeableness and extraversion and integrated it into a conversational agent. This serves as a novel approach to the traditional personality survey format (i.e., filling out a questionnaire). Adapting traditional questionnaire-type personality questions within the story allows users to provide self-assessments of their own personality while engaging with the conversational agent. We then used this information to examine the relationship between these aspects of a user's personality and their ratings. 

In this paper, we outline our reasoning and methods for developing and validating our story approach to the personality survey. Next, we explain how we implemented the personality story into a conversational agent. Finally, we discuss the results of the story method by first describing whether or not users engaged with the story and second by addressing our primary research question:  \textit{Does a user's personality play a role in the rating they provide?}

\section{Related Work}

Previous work in psychology has identified five traits that can be used to describe an individual's personality:  agreeableness, extraversion, neuroticism, openness, and conscientiousness \cite{goldberg1992bigfive}. Of these, extraversion and agreeableness are the two personality traits that have been shown to influence interpersonal communication most, as they point to characteristics such as sociability, affability and kindness \cite{mccraeCosta89, cupermanIckes09}. Examining human-human interactions, \cite{cupermanIckes09} demonstrate that individuals who score higher on extraversion are more likely to report their interaction as smooth, natural, and relaxed, and individuals who score higher on agreeableness as more likely to positively evaluate the quality of the interaction. 

Not only have extraversion and agreeableness been demonstrated to impact interpersonal communication between two human interlocutors, but these traits have also been shown to influence human-ai dyads.~\cite{astridKramerGratch10} examined how user personality traits influence their evaluations of interactions with text chatbots. The study consisted of participants completing a traditional personality survey and a short interaction with a chatbot who used the same five pre-recorded sentences with each participant. The results demonstrate that extraversion and agreeableness were better predictors of participants' interaction evaluations than the chatbot's behavior. 

While the importance of personality in human-ai conversations has influenced the development of personality classifiers \cite[e.g.,]{vinciarelli2014survey,ivanov2011recognition,mairesse2006automatic,mairesse2007using,rissola2019personality}, current automatic personality classification approaches were not suitable for our purposes. First, our experimental setup depends on text, not speech; and as such, we could not use a classifier that depends on spoken features \cite{ivanov2011recognition,mairesse2006automatic}. Second, our research question depends on an accurate assessment of personality in order to assess the relationship between personality and user ratings. Thus, we did not want to make the assumption that a classifier that is evaluated on a constructed dataset, a different demographic, or experimental setup in general \cite[e.g.]{mairesse2007using,rissola2019personality} would transfer to our human-ai interactions, where the expressions of personality are much more nuanced. Further, we would have no way of training or validating such models' performance, as due to several constraints, we could not ask users of our experimental agent to fill out a traditional personality survey.

\section{Method}

Taking into consideration that the personality survey would be implemented by a conversational agent, we determined that the traditional survey design (in the form of questionnaires) would not lead to engaging interactions. Instead, we developed a personality questionnaire in which the survey items are incorporated into a story that the conversational agent tells the user (see Table \ref{tab:story}). The conversational agent provides short scenarios that lead to questions for the user to engage with regarding their own self-assessment on the dimensions of extraversion and agreeableness. This method allows for a more natural conversational approach to eliciting personality questions. Further, by incorporating the survey items into a story, we expect users to feel more comfortable answering the questions, as the questions become part of the system sharing a story and attempting to create user engagement with the story. At the end of the story, the system asks the participant for feedback on the story as a means of understanding the performance of the story separate from the users' overall conversational experience. 

\begin{table}[]
    \centering
    \begin{tabular}{c|l}
        Turn & Conversational Agent Utterance \\
        \hline
        1 & I went to a fun party in the cloud yesterday, do you want to hear about it?\\
        & (User chooses whether or not to hear story) \\
        \hline
        2 & My friend was having a birthday party and I was so excited! \textbf{I just love going to parties,}\\
        & \textbf{what about you?} (extraversion) \\
        \hline
        3 & Oh interesting! Well, when I got on the bus, I saw my best friend! They’re my favorite \\
        & person to talk to! I can talk to them about anything. Umm, do you have someone in your \\
        & life like that?\\
        \hline
        4 &  Thanks for sharing. I think it’s because they do such a great job sympathizing with my \\
        & feelings. It usually helps me feel a lot better. \textbf{I'm just curious, are you good at} \\
        & \textbf{sympathizing with other people's feelings?} (agreeableness)\\
        \hline
        5 & I see, I'm not always very good at it. So when we got to the party, there were already lots \\
        & of people there! \textbf{Do you like talking to a lot of different people at parties?} (extraversion) \\
        \hline
        6 & When I was looking around, I saw someone who looked pretty sad. I thought we should go \\
        & check on them, but one of my friends wasn't interested. \textbf{Would you have gone to check}\\
        & \textbf{on them?} (agreeableness)\\
        \hline
        7 & Well, I decided to go over. It turns out they were having a really bad day, and they were \\
        & happy I came over to chat. We talked about their dog, and they showed me lots of pictures. \\
        & After a little while, I went back to find my friends. Would you like to hear more about the \\
        & party? (User chooses whether or not to continue story)\\
        \hline
        8 & It was so much fun! There was karaoke and dancing! Do you enjoy karaoke? \\
        \hline
        9 & Oh, interesting! I am not very good at it, but I love it! One of my friends didn't participate \\
        & too much, and kind of stayed in the background. \textbf{Do you tend to keep in the background}\\
        & \textbf{at parties too?} (extraversion)\\
        \hline
        10 & Yeah, I think it was because they didn't know that many people there. What about you?\\
        & \textbf{Are you usually quiet around new people?} (extraversion)\\
        \hline
        11 & Towards the end of the night, people brought out a big cake and sang happy birthday! It was \\
        & such a great night and party. I had a blast and met so many new people! Thanks for letting \\
        & me share my story. What did you think of it?
    \end{tabular}
    \caption{The story above was used to collect personality information about users. Not all turns collected personality information\textemdash some were used as simple engagement questions. The personality questions are in bold font, and the specific trait (extraversion or agreeableness) are identified in parentheses.}
    \label{tab:story}
\end{table}

This section serves a dual purpose. First, we will outline the method and results for developing and testing the personality story. Second, we will outline the method for implementing the story within the conversational agent. 

\subsection{Personality Story}
\label{sec-story}
In order to assess the quality of the results we could expect from our story method within the conversational agent, we first conducted a series of pilot studies in which we had participants complete a traditional personality survey \cite{goldberg2006ipip} shown in Table \ref{tab:survey}, our story survey, and five filler questions that asked participants to describe an image in five words or less. We semi-randomized the order:  participants either began the study completing the traditional survey or engaging with the story. The filler questions always occurred between these two in an effort to mitigate the repetition of the two personality-eliciting methods. 

\begin{table}[]
    \centering
    \begin{tabular}{l|l}
        Extraversion & Agreeableness \\
        \hline
        I am quiet around strangers & I feel others' emotions \\
        I start conversations & I am not really interested in others \\
        I don't like to draw attention to myself & I insult people \\
        I keep in the background & I have a soft heart \\
        I talk to a lot of different people at parties & I sympathize with others' feelings \\
        I have little to say & I take time out for others \\
        I don't mind being the center of attention & I make people feel at ease \\
        I don't talk a lot & I am not interested in other people's problems \\
    \end{tabular}
    \caption{For the traditional personality survey, users were presented with 16 statements and instructed to indicate how much the agreed or disagreed with each statement. The 16 items from \cite{goldberg99}'s Big Five Factor Markers for extraversion and agreeableness. }
    \label{tab:survey}
\end{table}

The traditional personality survey utilized a 6-point scale, wherein each numeric point represented the degree to which an individual agreed or disagreed with a given statement. Participant responses consisted of a numeric self-assessment of each survey item. For example, one survey item presented to participants was ``I am not interested in other people's problems.'' In contrast, the story method consisted of collecting text responses to the personality-probing questions embedded in the story. These questions were constructed to elicit yes/no responses from participants. For example, one excerpt from the story said ``I saw a person who looked pretty sad. I thought we should go check on them, but my friend wasn't really interested. Would you have gone over?'' These responses were then labelled using a 3-point scale (e.g., positive, negative, neutral). 

We collected data from approximately 100 participants on Amazon Mechanical Turk (mTurk). After excluding participants that did not follow the instructions, there were 96 participants for analysis. We analyzed the results of the story in two ways. First, we computed Cronbach's Alpha for the two traits (agreeableness and extraversion). Second, we calculated the mean score for each participant for agreeableness and extraversion for both the survey and the story, and then computed a simple linear regression to determine whether a participant's survey score could be predicted by their story score. 

\begin{table}
    \centering
    \begin{tabular}{c|c|c}
    \hline
     & Agreeableness & Extraversion\\ [0.5ex]
     \hline\hline
     Story & 0.42 & 0.9  \\
     \hline
     Survey &  0.84  & 0.91 \\
     \hline
     Story \& Survey & 0.86 & 0.94\\
     \hline
    \end{tabular}
       \caption{Cronbach's Alpha values for the story and survey on both personality traits.}
    \label{table:cronbachsalpha}
\end{table}

To see if an individual's overall score for agreeableness and extraversion from the story could predict their score on a traditional survey, we ran a simple linear regression. We found both of these models to be statistically significant (Agreeableness: R$^{2}$ = 0.3221,  F(1, 96) = 47.09, p $<$ 0.01; Extraversion:  R$^{2}$ = 0.6691, F(1, 96) = 197.1, p $<$ 0.01). 

These results suggest that for extraversion, the story items are both internally consistent \textbf{and} significant predictors for an individual's score on a traditional personality survey. For agreeableness, however, these results suggest that while the story may not be internally consistent, they still will predict an individual's score on a traditional personality survey. 

\subsection{Implementing the story in a conversational agent}

After validating that the story could replace a traditional personality survey, it was integrated into an experimental conversational agent that can have conversations on various topics such as movies, news, or pets. Users of the conversational agent are free to interact for as long as they like and at the end of each conversation, the conversational agent asks for feedback on a scale from 1 to 5. While the conversational agent is based on neural networks, the story was designed in a way that there is continuity regardless of the users' input, to avoid complicated flows that would make analyzing the results difficult. Therefore, it was implemented as a Finite State Machine, meaning that, for the story part, the conversational agent would move on to say pre-defined utterances ensuring that each user had the exact same experience. 

\begin{figure*}[t]
    \centering
    \includegraphics[width=0.8\textwidth]{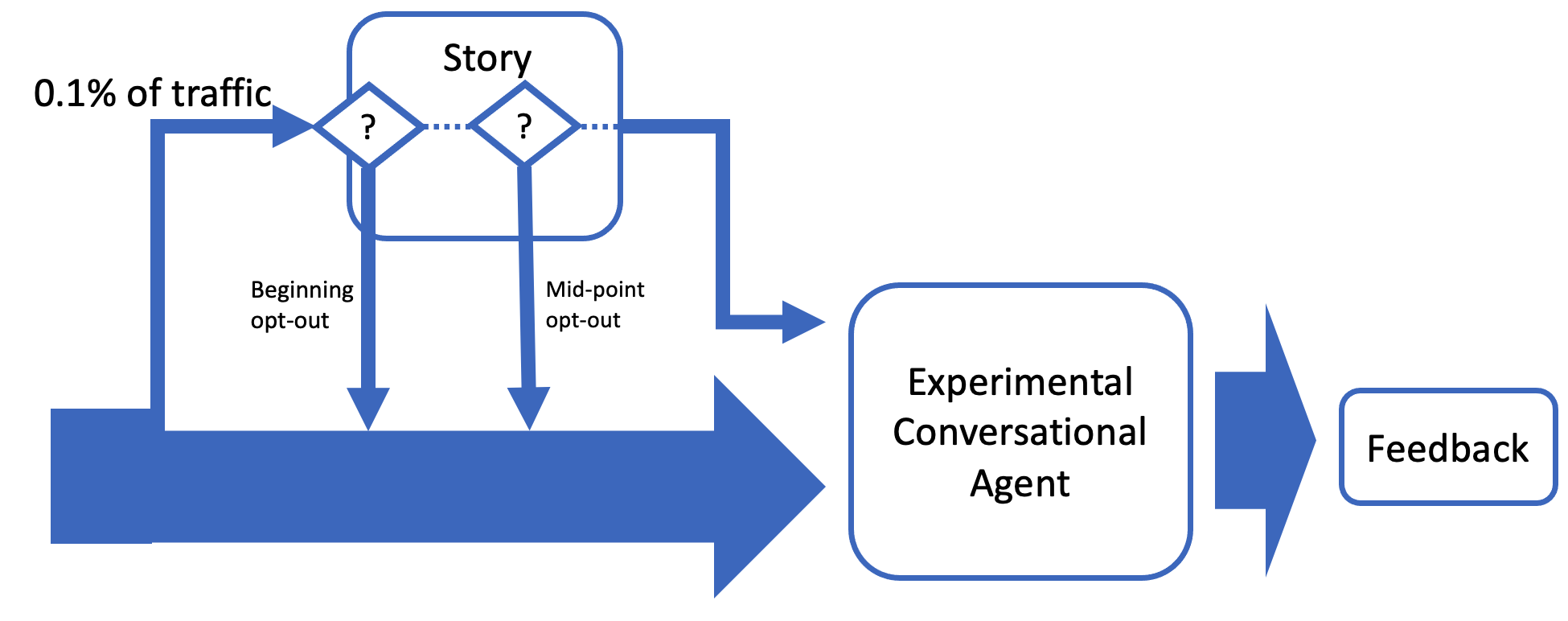}
    \caption{Flowchart showing how our story fits into the experimental conversational agent. `?' represents the beginning and mid-point question where users have the option to exit or continue with the story.}
    \label{fig:flow}
\end{figure*}

Our setup allows opt-in engagement. The conversational agent first asks the user if they want to hear a story and if the user responds positively, the story begins. Since the story is somewhat long (11 turns), we introduced a second point where the conversational agent re-affirms that the user wants to continue listening to the story. Figure \ref{fig:flow} shows the flow. Once the conversational agent completes the story, it asks the user what they would like to talk about next, thus allowing the user to continue the conversation as they wish. As such, while the story itself is FST-based, to ensure that each user heard exactly the same story, exactly the same way, the conversation after the story was dependent on the conversational agent.

\section{Results}

\subsection{User Reactions to Personality Story}
760 users were asked if they would like to hear the story. 70\% of them agreed and 30\% declined to hear the story. Users that engaged with the system's story include those that engaged with the entire story and those that engaged with part of the story. Although 760 users conversed with the agent, only 307 provided a rating for the overall conversation experience. Table \ref{table:ratingsByCategory} describes the distribution of those who engaged and did not engage with the story and shows that of these two groups, the average conversational score was higher for users who opted-in to the story. Welch's t-test showed the difference was statistically significant, t(113.16) = -3.3634, p = 0.001. Although it appears as though users who engage in the story have longer conversations than those who do not, the average number of turns in Table \ref{table:ratingsByCategory} include the story (11 turns). On average, users who listen to the story have the same length of conversation with the agent after the system tells its story as those who do not listen to the story. In other words, these users are not simply listening to the story and ending the conversation.

\begin{table}[h]
\setlength{\tabcolsep}{10pt} 
\centering
    \begin{tabular}{c | c c c }
        & Yes & No   \\
    \hline
    Total No. &532 & 228 \\
    \hline
    No. of Rated Conversations & 233 & 74 \\
    \hline
    Average Rating & 3.63 & 3  \\
    \hline
    Average No. of turns & 28.1 & 16.5\\
    \hline
    \end{tabular}
    \caption{Distribution of average rating and number of turns by those who engaged with the story (yes) and those who did not engage with the story (no).}
    \label{table:ratingsByCategory}
\end{table}

\subsection{Ratings through lens of User Personality}

\begin{figure}
    \centering
    \includegraphics[scale=0.3]{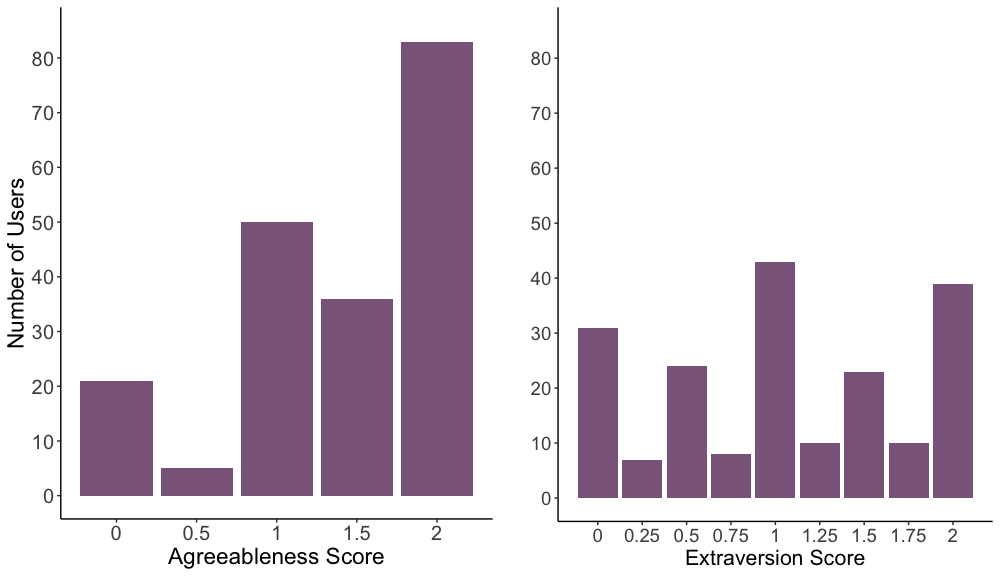}
    \caption{Number of users with a given average score for agreeableness (left) and extraversion (right). Note that that because there were a different number of questions for agreeableness (2) and extraversion (4) in the story (see Table \ref{tab:story}), there are different average scores for the two traits.}
    \label{fig:traitDistribution}
\end{figure}

Lastly, we address our original research question:  \textit{Does a user's personality play a role in the rating they provide?} To address this question, we examined the responses from 233 users who engaged with the personality story and provided a rating. We manually annotated the user responses to the personality questions. The responses are labelled on a 3-point scale (0 - 2). For example, if a user responds to the question ``Do you like going to parties, too?'' with ``No'', the user receives a 0 for extraversion. Through our manual annotation, we excluded participants whose responses could not be adequately interpreted, e.g. users who did not answer the questions or whose responses were not relevant. We then computed the mean scores for agreeableness and extraversion for 195 users, where a higher score indicates a higher level of agreeableness/extraversion. Figure \ref{fig:traitDistribution} shows the number of users with a particular agreeableness and extraversion average score. In general, we note that our sample is skewed for agreeableness -- users who score high on agreeableness were more likely to opt-in to the story and provide a rating. The scores for extraversion are more evenly distributed between high, mid, and low values.  

Next, we fit a general linear model using agreeableness and extraversion as predictors of conversational ratings. We found a significant relationship between agreeableness and conversational ratings, i.e., those who are more agreeable are more likely to provide a higher rating  ($\beta$ = 0.316, p = 0.02). However, the relationship between extraversion and conversational ratings is only nearing significance, i.e., those who are more extraverted are more likely to provide a higher rating ($\beta$ = 0.215, p = 0.098). Agreeableness and extraversion explained a small proportion of variance in conversational ratings, R$^{2}$ = 0.045, \textit{F}(2, 191)= 4.519, p = 0.012. 

For agreeableness, these results support the findings from \cite{astridKramerGratch10, cupermanIckes09}:  users who are more agreeable are more likely to rate their overall conversational experience higher than those who are not. For extraversion, however, more data is necessary in order to determine if the relationship is significant. It is important to note that the low R$^{2}$ value suggests that there are other factors that should be included to understand variation in ratings.

\section{Discussion}

In this study, we have constructed a conversational personality survey that can be implemented into a conversational agent that allows opt-in participation by means of engaging with a story. This approach provides two advantages. First, the story approach helps to mitigate the effects of social desirability bias \cite{edwards1957social, garrett2010attitudes} by focusing attention on engaging with a story. Second, utilizing an opt-in approach to the personality story, we avoid negatively impacting user experience. Not only is this approach advantageous for administering personality surveys, but it can be adapted to elicit other types of user characteristics that are typically obtained through traditional surveys. 


Results from the personality story suggest that extraversion does not predict users' overall experience with the experimental conversational agent. While this may be a reflection of a lack of data, this could also be a reflection in the difference between human-human interactions and human-ai interactions. Extraverted people are generally described as companionable, talkative, and confident \cite{cupermanIckes09, john1999big}, and the nature of the conversations between the user and the conversational agent are usually ones in which the conversational agent directs the conversation. In other words, a conversational partner of a conversational agent does not need to be confident and talkative, as the conversational agent tends to lead the conversation. 

On the other hand, results from the personality story show that agreeableness does predict overall conversation ratings. These results suggest that perhaps agreeableness between human-human interactions is more likely to transition to human-ai interactions than extraversion. Agreeable people are generally described as being sympathetic, cooperative, and considerate \cite{john1999big}. Sympathy and cooperation from the user can help to alleviate some of the conversational limitations of the agent. While it is possible that those who were more agreeable provided higher conversation ratings because the quality of the conversation was better, it is more likely that those who are more agreeable are simply more likely to rate conversations higher than those who are less agreeable. A qualitative review of the conversations post-story will need to be conducted in order to address this.

An interesting finding from this study is that users who chose to listen to the personality story tend to score high on agreeableness and also tend to provide higher ratings for their conversational experience. There are a few potential reasons that this could be the case. First, it is likely that users who are more agreeable are more likely to listen to the conversational agent's story. In this case, the results may be a case of selection bias:  users who are not agreeable opt-out of the story. Second, the story itself may be priming users' expectations of the conversational agent's capabilities. The story uses a fixed dialogue that does not adjust based on the user's response. These types of responses may very well lower a user's expectations of the conversational agent's capabilities. 

Taking these preliminary personality story results into consideration in conjunction with the differences in ratings based on Table \ref{table:ratingsByCategory}, it appears that users who agree to hear the story have a tendency to give a higher rating to the overall conversation than those who say no. Further, of those who agree to hear the story, the more agreeable a user is, the more likely they are to provide a higher rating for their conversational score. Future research needs to examine whether or not this alone (e.g. listening to a conversational agent's story) is sufficient to predict a user's rating and can be used to further personalize the conversational agent's interactions with users. 

\newpage

\bibliographystyle{plain}
\bibliography{mybib}

\newpage

\end{document}